\documentclass[letterpaper]{article} 
\usepackage{aaai2026}  
\usepackage{times}  
\usepackage{helvet}  
\usepackage{courier}  
\usepackage[hyphens]{url}  
\usepackage{graphicx} 
\usepackage{natbib}  
\usepackage{caption} 
\usepackage{algorithm}
\usepackage{algorithmic}
\usepackage{latexsym}
\usepackage{amssymb}
\usepackage{amsmath}
\usepackage{amsthm}
\usepackage{booktabs}
\usepackage{enumitem}
\usepackage{graphicx}
\usepackage{color}
\usepackage{multirow}
\usepackage{mathtools}
\usepackage{newfloat}
\usepackage{listings}
\DeclareCaptionStyle{ruled}{labelfont=normalfont,labelsep=colon,strut=off} 
\floatstyle{ruled}
\newfloat{listing}{tb}{lst}{}
\floatname{listing}{Listing}
\title{GIN-Graph: A Generative Interpretation Network for Model-Level Explanation of Graph Neural Networks}
\author {
    Xiao Yue\textsuperscript{\rm 1},
    Guangzhi Qu\textsuperscript{\rm 1},
    Lige Gan\textsuperscript{\rm 1}
}
\affiliations {
    \textsuperscript{\rm 1}Oakland University, Rochester, USA\\
    xiaoyue@oakland.edu, gqu@oakland.edu, lgan@oakland.edu
}

\usepackage{bibentry}

\begin{document}

\maketitle

\begin{abstract}
One significant challenge of exploiting Graph neural networks (GNNs) in real-life scenarios is that they are always treated as black boxes, therefore leading to the requirement of interpretability. To address this, model-level interpretation methods have been developed to explain what patterns maximize probability of predicting to a certain class. However, existing model-level interpretation methods pose several limitations such as generating invalid explanation graphs and lacking reliability. In this paper, we propose a new Generative Interpretation Network for Model-Level Explanation of Graph Neural Networks (GIN-Graph), to generate reliable and high-quality model-level explanation graphs. The implicit and likelihood-free generative adversarial networks are exploited to construct the explanation graphs which are similar to original graphs, meanwhile maximizing the prediction probability for a certain class by adopting a novel objective function for generator with dynamic loss weight scheme. Experimental results indicate that GIN-Graph can be applied to interpret GNNs trained on a variety of graph datasets and generate high-quality explanation graphs with high stability and reliability.
\end{abstract}
\section{Introduction}\label{sec1}
The increasing availability of graph data has led to the deployment of graph neural networks (GNNs) on various graph-related domains, including social analysis \cite{backstrom2011supervised}, biology, transportation, and financial systems \cite{wu2020comprehensive}. Inspired by neural networks which efficiently extract patterns from large and high-dimensional datasets, variants of GNNs \cite{micheli2009neural,scarselli2008graph} have been developed such as graph auto-encoders \cite{kipf2016variational}, graph recurrent neural networks \cite{li2015gated,tai2015improved}, graph attention networks \cite{velivckovic2017graph}, graph isomorphism networks \cite{xu2018powerful}, and graph convolutional networks. GNNs enable the application of machine learning techniques to graph-structured data. However, one significant challenge of exploiting GNNs in real-life scenarios is that they are typically treated as black boxes, which leads to the requirement for interpretability. In certain critical fields, the models can only be trusted if their predictions can be explained in human-understandable ways. In order to interpret and explain the underlying behaviors of neural networks when making predictions, several interpretation techniques have been proposed to explain deep learning models applied to image and text data \cite{simonyan2013deep,zhou2016learning}, which can be broadly categorized into two types: instance-level interpretation and model-level interpretation. Instance-level interpretations provide explanations for specific input instances by indicating the important features or the decision procedures for these inputs through the model. Model-level interpretations aim to interpret general behaviors of models by analyzing which patterns maximize the probability of predicting a certain class. One approach of interpreting GNNs at the instance level is to exploit the gradients or values in hidden feature maps as the approximations of input importance \cite{yuan2022explainability,li2022explainability}. Model-level interpretations on GNNs are less explored, as the input optimization method \cite{erhan2009visualizing} cannot be directly applied due to the discrete topological information of graphs like the adjacency matrix. Yuan et al. first proposed the XGNN \cite{yuan2020xgnn} interpreter to interpret GNNs at model-level. They trained a graph generator to create explanation graphs which maximize the prediction probability for a certain class. The graph generation is formulated as a reinforcement learning task with predefined policies. However, manual creation of these policies demands significant human effort. To address this limitation, GNNInterpreter (GNNI) \cite{wang2022gnninterpreter} was developed to build explanation graphs by learning a probabilistic generative graph distribution that produces the most discriminative graph pattern. However, two objective properties of generated explanation graphs that GNNI aims to maximize, prediction probability and embedding similarity, do not necessarily lead to a meaningful explanation graph. Moreover, it lacks an effective metric to evaluate the quality of explanation graphs, requiring manual effort to identify valid explanation graphs. In addition, the quality of generated explanation graphs is hindered by issues related to stability and reliability. To address limitations of GNNI, we proposed a new Generative Interpretation Network for Model-Level Explanation of Graph Neural Networks (GIN-Graph), to generate more reliable model-level explanation graphs. The implicit and likelihood-free generative adversarial networks (GANs) are exploited to construct explanation graphs which are topologically similar to real graphs, meanwhile maximizing the prediction probability for a certain class by adopting a novel objective function for generator with a dynamic loss weight scheme. The categorical reparameterization with the Gumbel-Softmax \cite{jang2016categorical} method is exploited to overcome the obstacle that a discrete adjacency matrix, which represents topological information of a graph, cannot be directly optimized via back-propagation. Besides, a graph pruning method is proposed to mitigate the challenge on generating fine-grained explanation graphs for models trained on dataset where the learned important patterns are much smaller than the full graphs, therefore enhancing the ability of GIN-Graph on capturing fine-grained topological features from the models. Experimental results indicate that GIN-Graph can be effectively applied on GNN models trained on various graph datasets and consistently create high-quality explanation graphs. The contributions of this paper are as follows:
\begin{itemize}
    \item We investigate the properties of model-level explanation graphs and define rules and a score metric for selecting valid explanation graphs.
    \item We develop the GIN-Graph, which incorporates a novel objective function for generator with a dynamic loss weighting scheme to create high-quality model-level explanation graphs for GNNs. A graph pruning preprocessing method is proposed to overcome the challenge in generating explanation graphs for models trained on dataset where the learned important patterns are much smaller than the full graphs.
\end{itemize}
It is important to clarify that explanation graphs discussed in this paper refer to model-level explanations for a GNN on a specific class, instead of explanations of a class within a graph dataset. A model-level explanation focuses on interpretation on the patterns learned by the model itself, therefore their generation must be guided by models' feedback. In contrast, explanations on a graph dataset aim to discover the pattern within the dataset, independent of any specific model. While two types of explanations may appear similar when a model has perfectly learned the dataset's patterns, they are fundamentally different in terms of their interpretive objectives.

\section{Related Works}
\subsection{GNNs Interpretations}
There is a rising interest in enhancing the interpretations of GNNs \cite{10.1145/3696444} due to the growing popularity of real-world graph data. One significant challenge of interpretations on graph-related models is the discrete nature of adjacency matrix that represents topology information of a graph, leading to difficulties in employing existing interpretation methods such as input optimization \cite{simonyan2013deep}. Based on the type of information that is being delivered, interpretations of GNNs are categorized as instance-level and model-level. Furthermore, instance-level interpretation methods can be classified into two categories: gradient-based and feature-based. Gradient-based methods compute the gradients of a target prediction relative to input features using back-propagation. Meanwhile, feature-based methods map the hidden features to the input space via interpolation to measure importance scores \cite{zhou2016learning}. As a gradient-based method, SA \cite{baldassarre2019explainability} directly exploits the squared values of gradients as the importance scores of different input features while importance scores can be directly computed via back-propagation. On the other hand, as a feature-based model, CAM \cite{pope2019explainability} measures important nodes by mapping the node features in the final layer back to the input space. However, it is limited to GNNs that employ a global average pooling layer and a fully-connected layer as the final classifier. GNNExplainer \cite{ying2019gnnexplainer} is a feature-based method that produces instance-level interpretation by learning a compact subgraph and a subset of node features that are most influential for a specific prediction. In contrast, model-level interpretations aim at explaining the behaviors of models. XGNN, which is the first model-level interpretation method on GNNs, exploits a graph generator to construct explanation graphs designed to maximize prediction probability for a certain class. This process involves iteratively adding an edge to a constructed graph, and utilizing feedback from the GNNs being explained to train the generator through policy gradient. And it can be formulated as a reinforcement learning task with predefined policies. However, creating these policies manually is labor-intensive. To address this limitation, GNNI learns a probabilistic generative graph distribution to generate the most discriminative graph pattern by optimizing a novel objective function, which is more flexible and computationally efficient compared to XGNN.

\subsection{Graph Generative Models}
Deep graph generative models have drawn significant attention due to the successful applications of generative models in other domains. Graph generative models share the common challenge in GNNs that machine learning techniques which are developed primarily for continuous data cannot be directly applied to a discrete adjacency matrix. To address this challenge, graph generative models are developed and can be categorized into two main branches \cite{guo2022systematic}, based on the generation process: sequential generation and one-shot generation. Sequential generation \cite{you2018graphrnn,zhang2019d} aims to construct nodes and edges in a sequential manner. In each step, one node and a few edges are generated. The edge generations involve predicting connections across all pairs of nodes or progressively selecting the nodes to be connected with the generated node from the existing nodes. In contrast, one-shot generation \cite{flam2020graph,bresson2019two} exploits a probabilistic graph model based on the matrix representation which allows the simultaneous generation of all nodes and edges. These models learn to encode graphs into latent representations based on a probabilistic distribution, therefore a graph can be obtained by directly sampling from it in a single step. Graph generative models are extensively utilized in graph-centric fields such as molecule design \cite{de2018molgan,popova2019molecularrnn} and protein structure modeling \cite{anand2018generative}.

\section{Model-level explanation graphs}
\label{Model-level explanation graphs}
\subsection{Valid and invalid explanations}
\label{valid or invalid}
\begin{figure}[htb]
    \centering
    \includegraphics[scale=0.48]{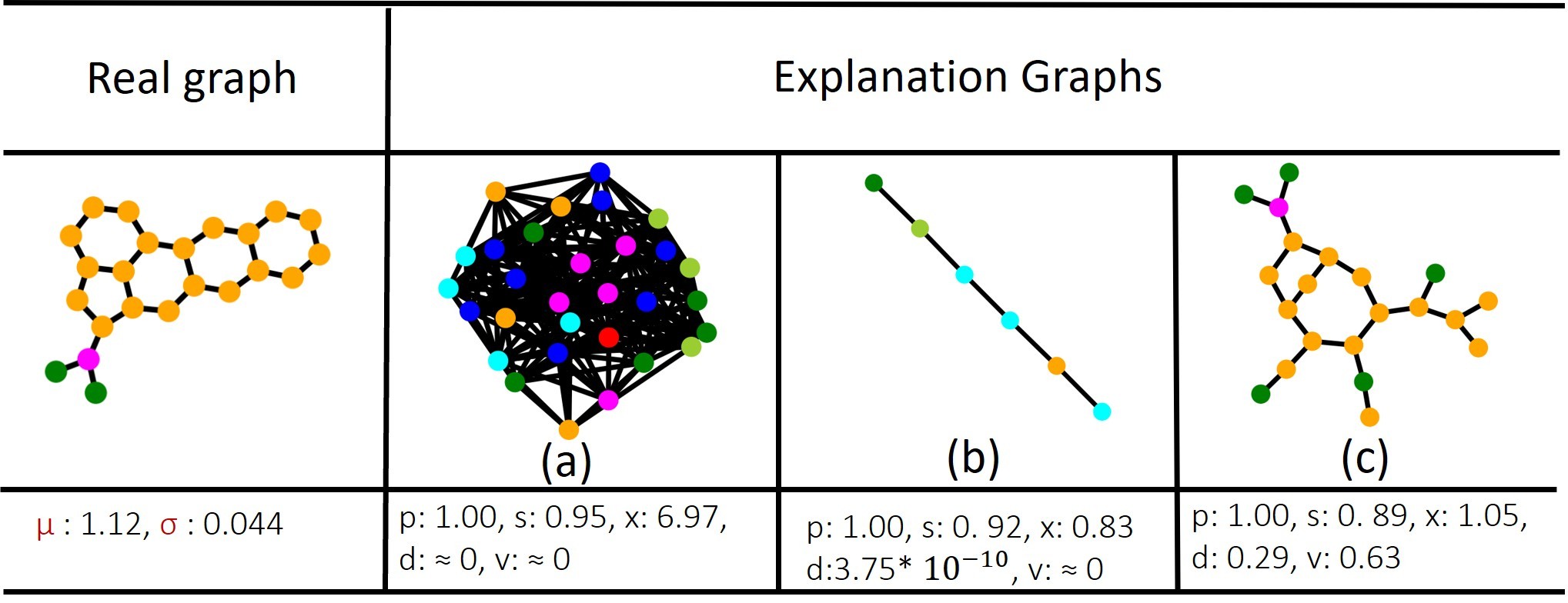}
    \caption{Illustration of valid and invalid explanation graphs}
    \label{examplefake}
\end{figure}

Model-level interpretations of GNNs aim to provide explanation graphs that maximize the probability of predicting a specific class. Intuitively, one might consider explanation graphs that the GNN model predicts with high probability as valid explanation graphs. However, it is insufficient to only rely on the probability given by GNNs to evaluate explanation graphs. Note that most GNNs are not trained to directly handle out-of-distribution (OOD) graphs, therefore some graphs with high probability may be OOD graphs and do not provide any meaningful information to the user. GNNI addresses this issue by maximizing the similarity between the embedding of an explanation graph and the average embedding of all graphs of a certain class. Nonetheless, these two objectives, prediction probability and embedding similarity, are heavily dataset-dependent and may still lead to an invalid explanation graph. We show counter examples in Figure \ref{examplefake} to illustrate the issue. In this context, there is a GNN trained on the \emph{MUTAG} dataset. Explanation graphs (a) and (b) achieve nearly 100\% prediction probabilities and exhibit cosine similarities above 0.9 in their embeddings for class \emph{Mutagen}. However, unlike explanation graph (c), they fail to provide meaningful insights for model-level interpretation. Therefore, we consider the explanation graphs (a) and (b) are invalid, and graph (c) is a valid one. This indicates that relying solely on prediction probability and embedding similarity to assess explanation quality may introduce bias. 
GNNI attempts to eliminate some invalid explanation graphs by limiting the maximum number of edges in a graph through a hyperparameter, which requires manual fine-tuning due to varying number of nodes across different graphs. However, this method is insufficient to filter out other invalid graphs, such as explanation graph (b). To overcome this limitation, we further utilize the average degree $x$ of a graph ($x = e/n$, where $e$ is the number of edges and $n$ is the number of nodes) to evaluate explanation graphs. We compute the mean ($\mu$) and standard deviation ($\sigma$) of the average degrees of all graphs in a certain class. Any graphs whose average degrees fall outside the range $[\mu-t*\sigma,\mu+t*\sigma]$ are considered invalid explanation graphs, with $t=3$ as a commonly used threshold. As shown in Figure \ref{examplefake}, explanation graphs (a) and (b) are considered invalid, as their average degrees fall outside of the acceptable range.
Additionally, to more effectively evaluate the explanation graphs, we define a validation score $v$ as $v=\sqrt[3]{s*p*d}$ which integrates embedding similarity ($s$), prediction probability ($p$) and degree score ($d$), where the degree score is calculated as $d=e^{-\frac{(x - \mu)^2}{2\sigma^2}}$, and $x$ is the average degree of an explanation graph. In this work, we use validation score to assess the quality of an explanation graph. The score ranges from \emph{0} to \emph{1}, with \emph{1} representing the highest quality. It is designed to be sensitive to low values, ensuring that an explanation graph cannot achieve a high score if it performs poorly in any one aspect. 
\subsection{Fine-grained and coarse-grained explanations}
\begin{figure}[htb]
    \centering
    \includegraphics[scale=0.37]{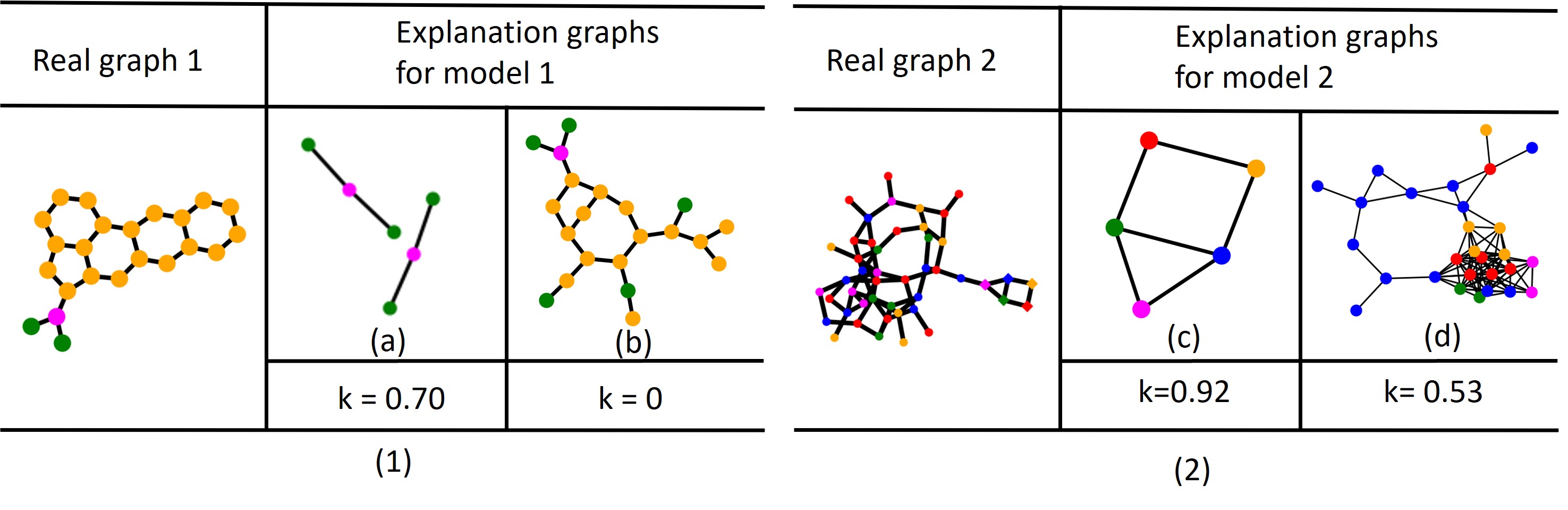}
    \caption{Illustration of fine-grained and coarse-grained explanation graphs}
    \label{local}
\end{figure}

Explanation graphs can vary significantly in an important property: size, even though they all satisfy the requirements for being valid. A smaller explanation graph tends to highlight the fine details of a model-level explanation, capturing detailed patterns the model has identified as significant. In contrast, a larger explanation graph typically tends to present the broader structural patterns, reflecting the high-level relationships the model has learned. Accordingly, explanation graphs can be further categorized into two types based on their level of detail: \emph{fine-grained} and \emph{coarse-grained}. Figure \ref{local} shows fine-grained and coarse-grained explanation graphs of two models learned trained on two datasets : \emph{MUTAG} and \emph{motif}. Fine-grained explanation graphs have high granularity, such as explanation graphs (a) and (c), representing important detailed patterns learned by GNNs. For instance, explanation graph (a) demonstrates the $NO_{2}$ structure learned by the model which is trained on the \emph{MUTAG} dataset, while explanation graph (c) presents a motif, which is identified as an important pattern by the model trained on the \emph{motif} dataset. In contrast, coarse-grained explanation graphs with low granularity, such as explanation graphs (b) and (d), tend to be similar to original graphs to present broader structural patterns. In this work, we define granularity related explanation metric ($k$) using the formula $k=1 - min(1,b/a)$, where $a$ is the average number of nodes across all graphs in a certain class and $b$ denotes the number of nodes in an explanation graph for that class. This formula scales the granularity into the range of \emph{0} to \emph{1}, where \emph{0} indicates the lowest granularity and \emph{1} represents highest granularity which is impossible in practice. 
\section{Approaches}
\begin{figure}[!htb]
    \centering
    \includegraphics[scale=0.54]{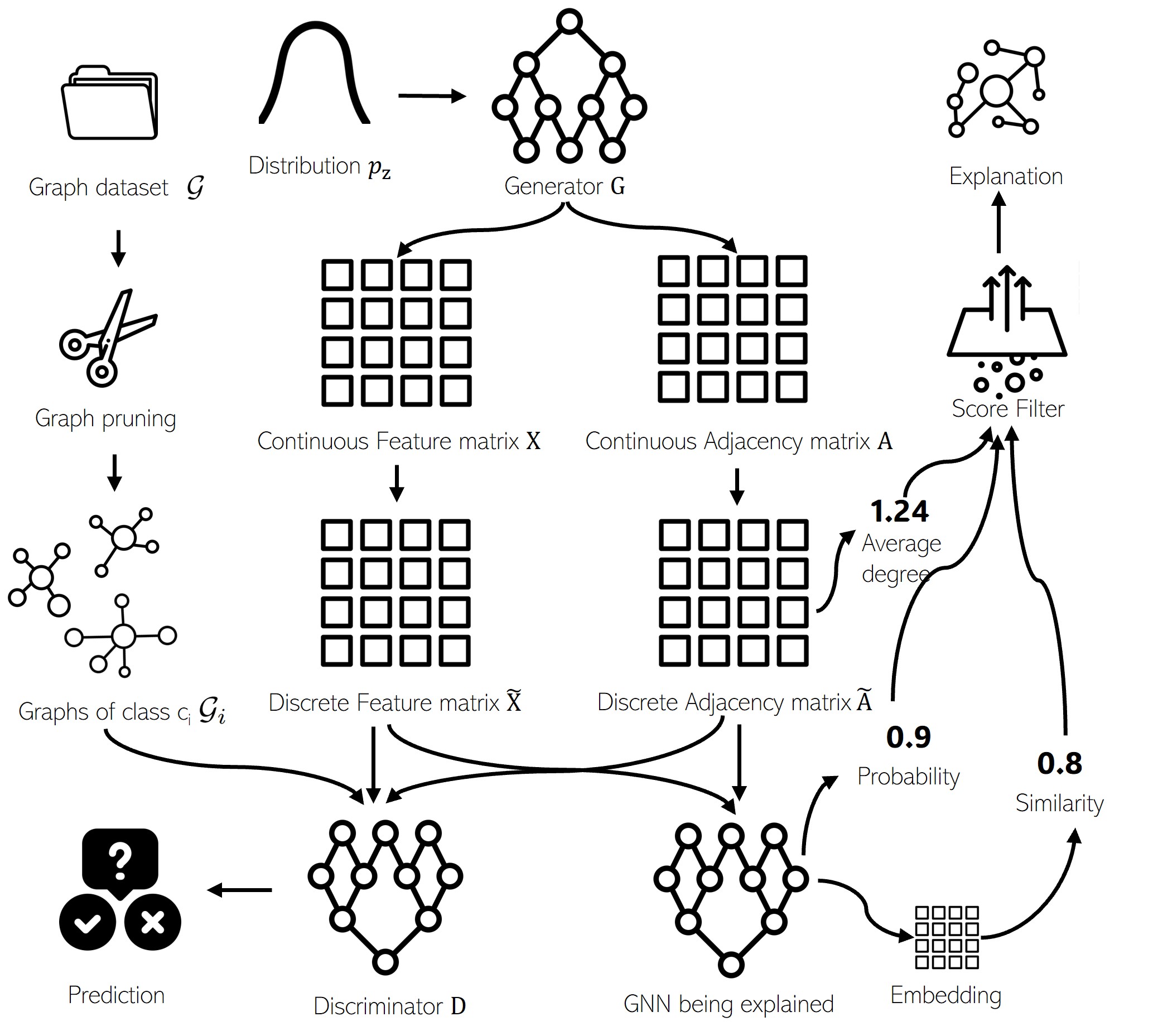}
    \caption{Overview of the Architecture of GIN-Graph}
    \label{GIN-Graph}
\end{figure}

We present details of the GIN-Graph in this section. The overview of GIN-Graph is shown in Figure \ref{GIN-Graph}. It exploits a generative adversarial network to construct explanation graphs, using the feedback from the GNN to be explained. 

\subsection{Optimization goal}
Let $f(\cdot)$ denote a GNN model that we aim to interpret, which is trained on a graph dataset $\mathcal{G}$ with classes of $\{c_{1}, ... , c_{l}\}$. Hence, for a target class $c_{i}$, the model-level interpretation of $f(\cdot)$ on class $c_{i}$ can be formulated as deriving an explanation graph $g*$ such that $g* = \arg \max_{g} P(f(g)=c_{i})$. Since the prediction probability should not be the only metric for assessing the quality of an explanation graph, we introduce a score filter denoted as $W(\cdot)$, which takes a graph as input and determines if the graph satisfies the preset rules, such as the validation score ($v$) is larger than a specific threshold. Considering the validation metric, the objective of generating an explanation graph is building a graph $g*$ such that $g* = \arg \max_{g} P(f(g)=c_{i}|W(g)=True)$.

\subsection{Generative Adversarial Network}
\subsubsection{Generator and Discriminator}
 We denote the set of all graphs in class $c_{i}$ as $\mathcal{G}_{i}$. A generative adversarial network consisting of a generator $G(\cdot)$ and a discriminator $D(\cdot)$, is exploited to generate explanation graphs, while the parameters of $G(\cdot)$ and $D(\cdot)$ are denoted as $\theta_{G}$ and $\theta_{D}$, respectively. The generator $G(\cdot)$ is implemented as a multi-layer perceptron, which takes a sampled vector $z$ from a distribution $p_{z}$ as input and generates graphs, denoted as $G(z) = \hat{g}$. In this case, two matrices with continuous values, an adjacency matrix and a node feature matrix, represent the topology of a graph for downstream tasks. The parameters of the generator $\theta_{G}$ are optimized using the objective function defined in Equation \ref{loss}, where $L_{GAN}=-\mathbf{E}_{z\sim p_{z}}[D(G(z))]$ is the original GAN loss function and $L_{GNN}$ is the cross-entropy loss or logits loss.
\begin{equation}
\label{loss}
    \theta_{G} = \arg \min_{\theta} ((1-\lambda)L_{GAN} + \lambda L_{GNN}(f(\hat{g}),c_{i})) 
\end{equation}
As both generated and real graphs are represented by an adjacency matrix and a node feature matrix, we exploit a GNN as the discriminator for taking graphs as inputs. The discriminator $D(\cdot)$ is utilized for differentiating a real graph from a generated one. We denote the distribution of existing graphs as $p_{x}$. Parameters of the discriminator are optimized by Wasserstein GAN (WGAN) loss with gradient penalty shown in Equation \ref{loss_discriminator}, where $\hat{x} = \epsilon x + (1-\epsilon)G(z), \epsilon \sim \mathcal{U}(0,1)$ and $\alpha$ is a hyperparameter of gradient penalty. 
\begin{equation}
\begin{split}
\label{loss_discriminator}
    \theta_{D} = \arg \min_{\theta} \mathbf{E}_{z\sim p_{z}}[D(G(z))] - \mathbf{E}_{x\sim p_{x}}[D(x)] +\\ \alpha(\lVert{\nabla_{\hat{x}}D(\hat{x})}\lVert-1)^{2}
\end{split}
\end{equation}
\subsubsection{Adjacency matrix and node feature matrix}
The generator $G(\cdot)$ outputs two matrices with continuous values, a continuous adjacency matrix $A$ and a continuous node feature matrix $X$. However, if there is no attribute associated with edges in a graph, each element in the discrete adjacency $\widetilde{A}$ should be either \emph{0} or \emph{1}, indicating the absence or presence of an edge, respectively. The discrete adjacency matrix can be derived through binary categorical sampling from $A$ in this scenario. If edges have categorical features with \emph{k} different classes, these features are encoded into a (\emph{k+1})-dimensional vector by one-hot encoding, where the extra dimension indicates the absence of an edge. In this case, we treat each element in the adjacency matrix as a (\emph{k+1})-dimensional vector, instead of a single value. The node feature matrix $X$ can be directly feed into the discriminator for prediction if node features are numerical. If the node features are categorical, we exploit the same categorical sampling method used for adjacency matrix to derive a discrete node feature matrix $\widetilde{X}$. Note that the categorical sampling operation is not differentiable, therefore cannot be directly optimized by back-propagation. We address this obstacle by exploiting the categorical reparameterization with the Gumbel-Softmax \cite{jang2016categorical}.
\subsubsection{GNN being explained}
Similar to the discriminator $D(\cdot)$, the GNN that is being explained should also takes graph data as inputs. The GNN outputs embedding and prediction probability of a generated explanation graph for evaluation. Note that the GNN being explained doesn't have to be the same model structure as the discriminator, as long as it is capable of accepting an adjacency matrix and a node feature matrix as inputs, or any other data formats that can be derived through differentiable transformations from these two matrices.
\subsection{Dynamic loss weight scheme}
\subsubsection{Pitfalls of gradients from GNNs}
One might envision an ideal model-level GNN interpreter as a model which takes a pre-trained GNN as input, and generates explanation graphs that maximize the prediction probability of a certain class without relying on any graphs from the original training dataset. However, as most GNNs are not trained to directly handle out-of-distribution graphs, which most are invalid explanation graphs, the gradients derived by feeding such graphs into the GNNs do not necessarily guide the generator toward producing valid explanation graphs. As a result, the generator tends to converge toward generating invalid explanation graphs with maximum prediction probability. We conducted several experiments where a generator was trained by solely optimizing the prediction probability from a GNN. The experimental results consistently showed that the generator quickly converged to generating invalid explanation graphs. This suggests that training a generative model to produce valid graph explanations solely through backpropagation from the GNN is highly challenging especially when the model is randomly initialized, therefore highlighting the necessity of using the original training dataset for interpretation.
\subsubsection{Time-dependent weighting parameter}
Compared to the large and unstable gradients derived from the GNN, original GAN loss derived from the discriminator typically produces smaller and more stable gradients for the generator due to the adversarial training strategy. Training the GAN model on original dataset solely by GAN loss makes the generator tend to produce graphs that are similar to those in the training dataset, rather than graphs that maximize the prediction probability of a certain class. However, directly combining the GAN loss ($L_{GAN}$) with the loss derived from the GNN that is being explained ($L_{GNN}$) can introduce a potential risk: the $L_{GNN}$ may dominate the training process, leading the generator to converge into generating invalid explanation graphs. To mitigate this risk, we proposed a dynamic weighting scheme by developing a time-dependent weighting function for the $\lambda$ parameter in Equation \ref{loss}. The weighting function is shown as Equation \ref{dynamic loss}, where $\lambda_{\min}$ and $\lambda_{\max}$ denote the minimum and maximum values of the weight parameter; $t$ is the current training iteration, and $T$ is the total number of iterations. Ratio of the training period to increase $\lambda$ is defined by $p$, and $k$ is a temperature parameter for adjusting the sharpness of the transition. Function $\sigma$ represents a weight adjustment functions such as a \emph{Sigmod} function. This dynamic weighting scheme ensures that the $L_{GAN}$ dominates the early stages of training to guide the generator for producing graphs that are similar to those in the training dataset, while $L_{GNN}$ gradually takes effect in the later stages to fine-tune the explanation graphs for maximum prediction probability.
\begin{equation}
\label{dynamic loss}
\lambda(t) =\lambda_{\min} + \bigl(\lambda_{\max} - \lambda_{\min}\bigr) \cdot 
\sigma\!\left(k \left( 2 \cdot \dfrac{\tfrac{t}{T} - p}{1 - p} - 1 \right)\right)
\end{equation}

\subsection{Graph pruning}
\label{pruning}
Most of message-passing based GNNs utilize the important patterns (e.g., subgraphs) in predictions \cite{ying2019gnnexplainer}, implying that some nodes and edges may be redundant and contribute little or nothing to the prediction. In order to improve the explanation graphs for GNNs that are trained on datasets where important patterns are much smaller than the entire graphs, GIN-Graph employs a graph pruning method that preprocesses the input graphs by removing nodes and edges that are irrelevant to prediction. This method enhances the capability of GIN-Graph in generating fine-grained explanation graphs of GNNs. The graph pruning process iteratively removes nodes and edges in a graph at random, and each resulting graphs is evaluated by the GNN that being explained: if the prediction probability does not drop, the change is retained; otherwise it is discarded.

\section{Experiments}
\subsection{Dataset and experiment setting}
GIN-Graph was evaluated on four GNN models trained on four datasets: one real-world dataset \emph{MUTAG} and three synthetic datasets \emph{Cyclicity}, \emph{Shape} and \emph{Motif}. The \emph{MUTAG} \cite{debnath1991structure} dataset is a widely recognized benchmark dataset in the field of graph learning. Each compound sample is labeled according to its mutagenic effect on a bacterium. The datasets \emph{Cyclicity}, \emph{Shape} and \emph{Motif} are introduced in the paper \cite{wang2022gnninterpreter}, and they are randomly generated by their respective algorithms. 
GIN-Graph is implemented using PyTorch with python version 3.10. In order to compare GIN-Graph to GNNI, the state-of-the-art model-level GNN interpreter, we evaluated both interpreters on same GNNs. Since GNNI has already been shown to outperform XGNN, we focus on comparing GIN-graph to GNNI in this paper. We applied the graph pruning method to \emph{Cyclicity} and \emph{Motif} datasets to derive smaller graphs as inputs when generating explanation graphs for models trained on these two datasets. Note that as GNNI doesn't take graphs as inputs, the graph pruning method cannot be applied to GNNI. We trained GNNI 100 times for each class of a GNN and selected the best explanation graphs from all generated graphs based on observation and prior knowledge of the datasets. We also conducted experiments on training GIN-Graph 100 times of each class and selected best explanation graphs as results using the same criterion applied to GNNI.
\subsection{Experimental results}
\begin{figure*}[htb]
    \centering
    \includegraphics[scale=0.47]{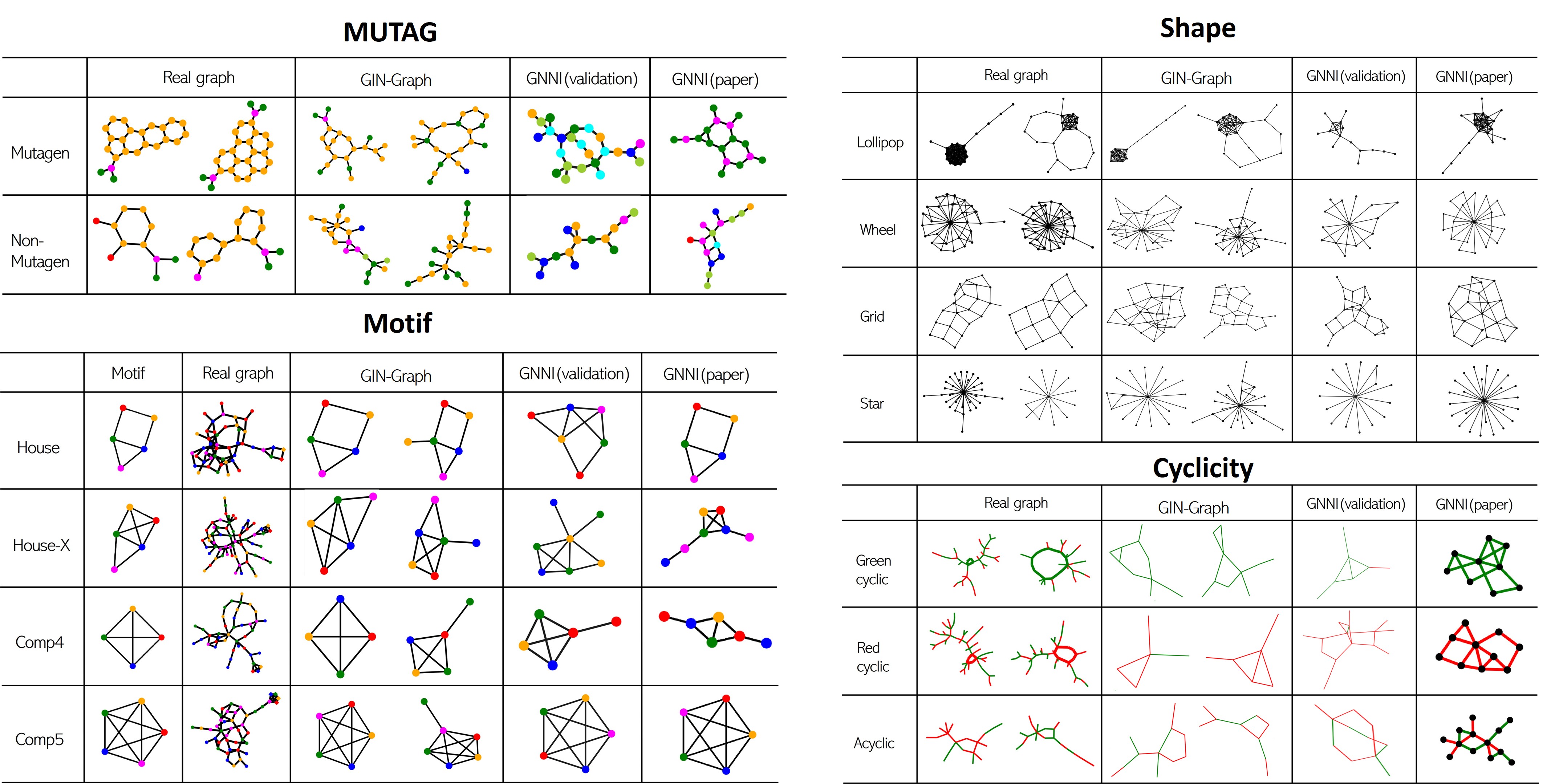}
    \caption{Experimental results on GNNs trained on four datasets}
    \label{experiments}
\end{figure*}

The experimental results for GNNs trained on datasets \emph{MUTAG}, \emph{Motif}, \emph{Shape} and \emph{Cyclicity} are shown in Figure \ref{experiments}.
The best explanation graphs of each class generated by GNNI are presented in the column \textbf{GNNI (validation)}. However, due to the loss of original data and the lack of detailed guidance on hyperparameter settings, we were unable to reproduce all results in the GNNI paper, despite conducting extensive experiments with ample training time. Therefore we also present the explanation graphs from their paper \cite{wang2022gnninterpreter} in column \textbf{GNNI (paper)} for reference. Experimental results indicate that GIN-Graph demonstrates superior performance to GNNI, as GIN-Graph generated high quality explanation graphs for all classes. For the GNN trained on \emph{Motif} dataset, GIN-Graph accurately generated all motifs that GNN learned while GNNI failed for classes \emph{house}, \emph{house-X} and \emph{comp4}. For the GNN trained on \emph{MUTAG} dataset, neither interpreter could present the fact that both $NO_{2}$ and $NH_{2}$ substructures are strong evidences for the molecule mutagenicity \cite{debnath1991structure}. However, GIN-Graph was able to capture the carbon-based chemical compound structures of molecules, and the number of carbon atoms is much larger than that of other atoms. For the GNNs trained on \emph{Shape} and \emph{Cyclicity} datasets, both interpreters produced valid explanation graphs for all classes.
However, evaluating a model’s performance only based on the best explanation graph among hundreds of graphs is insufficient. Therefore, to more comprehensively evaluate the overall quality of generated explanation graphs, we trained each model with selected best explanation graphs' parameter settings for 10 runs on each class. For each run, we selected the top 10 explanation graphs based on validation scores, resulting in a total of 100 graphs for each class per model. To ensure a fair comparison on two models, we disabled the filter mechanism in GIN-Graph, so that low-scoring graphs were not excluded to get advantages. We computed the average validation scores of selected explanation graphs and present the experimental results on Table \ref{score}. Note that objective function of GNNI optimizes both prediction probability and embedding similarity, two measurements in the validation score, while prediction probability is the only measurement being optimized in the objective function of GIN-graph. From this perspective, the validation score metric inherently favors GNNI. Despite this disadvantage, GIN-Graph consistently generated explanation graphs with higher average validation scores than GNNI on all classes, demonstrating its effectiveness in generating high-quality explanations.

\begin{table}[htb]
\caption{Experimental results on validation scores of models trained on four datasets}
\label{score}
\small
\begin{tabular}{clll}
\hline
\multirow{2}{*}{Dataset}   & \multicolumn{1}{c}{\multirow{2}{*}{Class}} & \multicolumn{1}{c}{\multirow{2}{*}{GIN-Graph}} & \multicolumn{1}{c}{\multirow{2}{*}{GNNI}} \\
                           & \multicolumn{1}{c}{}                       & \multicolumn{1}{c}{}                           & \multicolumn{1}{c}{}                      \\ \hline
\multirow{2}{*}{MUTAG}     & Mutagen                                    & 0.972                                          & 0.904                                     \\
                           & Non-Mutagen                                & 0.981                                          & 0.935                                     \\ \hline
\multirow{4}{*}{Shape}     & Lollipop                                   & 0.994                                          & 0.781                                     \\
                           & Wheel                                      & 0.985                                          & 0.923                                     \\
                           & Grid                                       & 0.994                                          & 0.916                                     \\
                           & Star                                       & 0.999                                          & 0.935                                     \\ \hline
\multirow{4}{*}{Motif}     & House                                      & 0.927                                          & 0.157                                     \\
                           & House-X                                    & 0.910                                          & 0.106                                     \\
                           & Comp4                                      & 0.955                                          & 0.187                                     \\
                           & Comp5                                      & 0.846                                          & 0.578                                     \\ \hline
\multirow{3}{*}{Cyclicity} & Red Cyclic                                 & 0.989                                          & 0.954                                     \\
                           & Green Cyclic                               & 0.999                                          & 0.847                              \\
                           & Acyclic                                    & 0.965                                         &  0.886                             \\ \hline
\end{tabular}
\end{table}
\subsection{Experiments on convergence states and stability}
\begin{figure}[htb]
    \centering
    \includegraphics[width=0.95\linewidth]{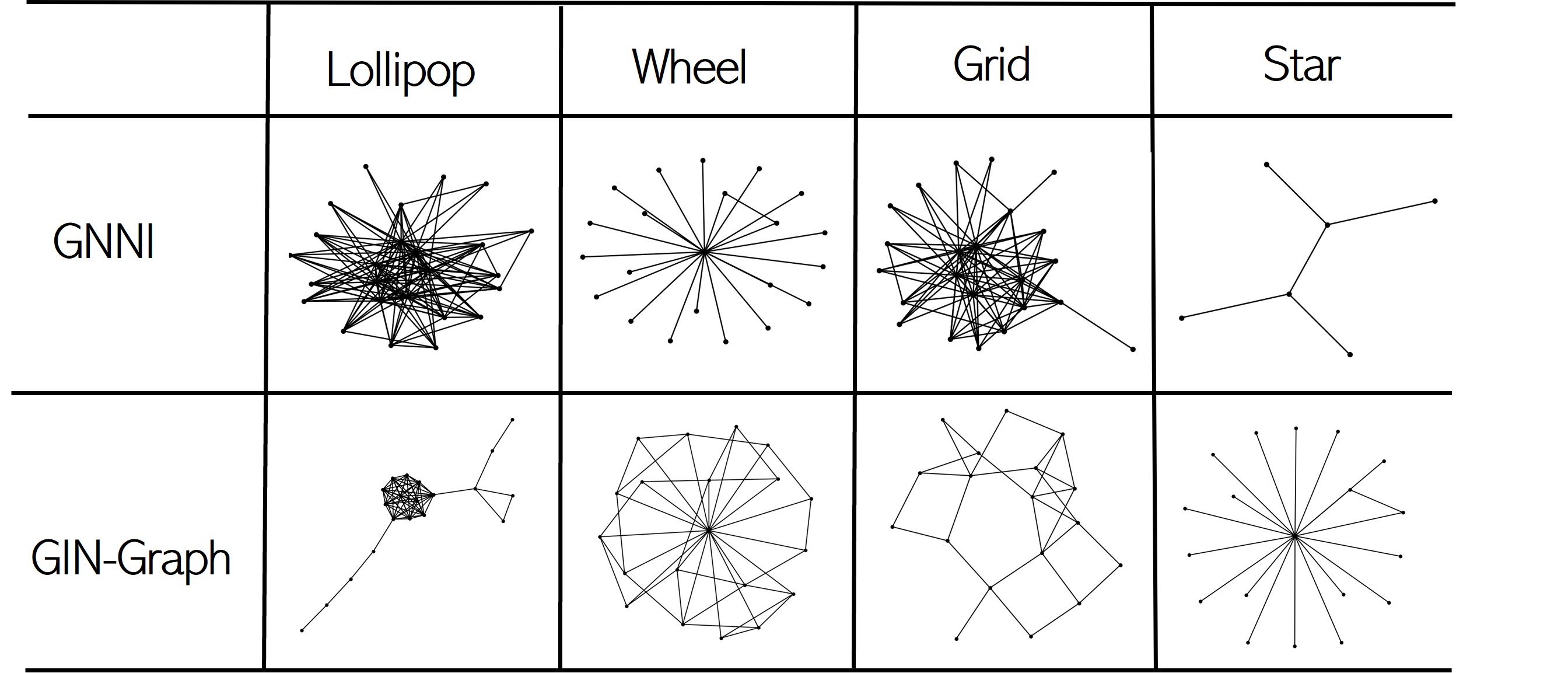}
    \caption{Generated explanation graphs after convergence}
    \label{converge}
\end{figure}

During our experiments, we observed that all valid explanation graphs from GNNI were generated at the early stages of training, typically within first 100 iterations. If no valid explanation graphs were generated at the early stages, continued training GNNI did not lead to their emergence. In contrast, GIN-Graph was able to consistently generated valid explanation graphs with sufficient training time. This contrast leads to investigation on convergence states of the two models. We select the GNN trained on \emph{Shape} dataset to be explained for this analysis, as each class in this dataset exhibits distinct topological features. We trained two models with extended training time and monitored the generated graphs to evaluate the convergence states. We treated the convergence as the state where generated graphs remained stable over a long period. Each model was trained 10 times per class and the results were consistent across runs, as shown in Figure \ref{converge}. Experimental results indicate that convergence states of GNNI do not consistently correspond to valid explanation graphs. In contrast, GIN-Graph consistently converged at states where high-quality valid explanations are generated. Our findings also reveal a key weakness of GNNI: its instability. In experiments, GIN-Graph successfully generated valid explanation graphs in all 10 runs for every class, while GNNI only succeeded in 2, 9, 4, and 10 out of 10 runs for the \emph{Lollipop}, \emph{Wheel}, \emph{Grid}, and \emph{Star}, respectively. Even though with identical hyperparameters, GNNI frequently failed to generate valid explanation graphs for classes \emph{Lollipop} and \emph{Grid} due to incorrect convergence states, suggesting that valid explanation graphs are produced because of randomness during the training process. Experimental results emphasize the low stability and limited reliability of the GNNI model. 
\subsection{Experiments on effectiveness of graph pruning}
\begin{figure}[htb]
    \centering
    \includegraphics[scale=0.25]{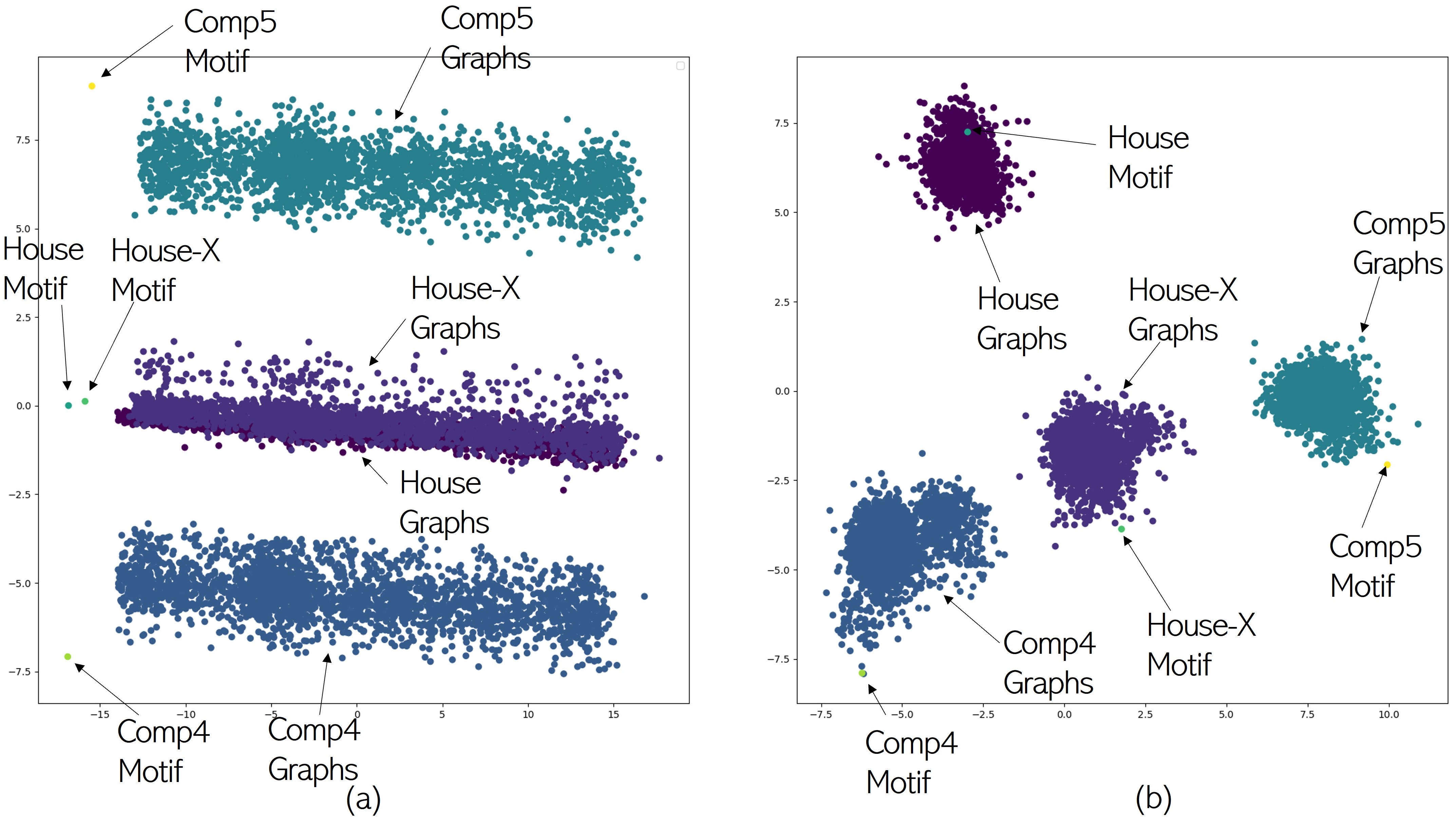}
    \caption{Embedding distributions of graphs before (a) and after (b) graph pruning}
    \label{embed}
\end{figure}

In the section \emph{Graph pruning}, we proposed the graph pruning preprocessing method designed to remove nodes and edges that are irrelevant to predictions by a GNN model. We used the \emph{Motif} dataset and the GNN trained on it to evaluate the effectiveness of this method. Each graph in the \emph{Motif} dataset is constructed by connecting a random Rome graph to a specific motif. There are four different motifs: \emph{house}, \emph{house-X}, \emph{comp4}, and \emph{comp5}, as shown in Figure \ref{experiments}. In this dataset, the type of motif is the only distinguishing factor between different classes. We first evaluated the prediction accuracies of the GNN model on original graphs. Then, we assessed the accuracy of the same GNN model on the same graphs with all motifs removed. For graphs with motifs, the prediction accuracies of \emph{house}, \emph{house-X}, \emph{comp4}, and \emph{comp5} were 99.21\%, 99.80\%, 99.13\%, and 100.00\%, respectively. For graphs without motifs, the prediction accuracies of them were 0.17\%, 0.00\%, 0.00\%, and 0.00\%, respectively. Therefore we can conclude that motifs are the only critical patterns for the GNN to make accurate predictions. However, as the number of nodes in a motif (4 or 5 nodes) is extremely smaller than the number of nodes in a graph (more than 50 nodes), we applied the graph pruning method to remove redundant nodes and edges. We observed that all motifs remain intact regardless of the pruning intensity. To investigate the relationships between the motifs and graphs, we projected the embeddings of the graphs obtained from the GNN into two-dimensional space, along with the embeddings of the motifs. We observed that none of the motifs are located within the clusters of original graph embeddings, indicating that they are outliers, as shown in Figure \ref{embed} (a). This suggests that generating fine-grained explanation graphs that are exactly the same as motifs by exploiting embeddings is challenging. This observation aligns with our experimental results, where GNNI cannot construct the motifs for classes \emph{house}, \emph{house-X} and \emph{comp4}. We also generated distributions of the embeddings of the pruned graphs, as shown in Figure \ref{embed} (b). Comparing to Figure \ref{embed} (a), we observed that the embeddings of pruned graphs are more tightly clustered and the embeddings of \emph{House} and \emph{House-X} graphs are well-separated. Additionally, the embeddings of the motifs are closer to their respective clusters, indicating that the pruned graphs are more similar to the real motifs. 
However, as the graph pruning was originally designed to simplify graphs whose important patterns are much smaller than the graphs themselves, it is risky to apply graph pruning on graphs whose important patterns learned by GNNs are less clear. After evaluating the effectiveness of graph pruning on the \emph{MUTAG} dataset, we found the pruned graphs are more likely to retain the $NO_{2}$ structure. Considering the fact that both $NO_{2}$ and $NH_{2}$ substructures are strong evidence for the molecule mutagenicity, we could draw the conclusion that the GNN model learned the important patterns partially. However, graph pruning may break the carbon-based structure, such as carbon rings, resulting in explanation graphs that no longer resemble valid chemical compounds. In conclusion, graph pruning is effective in deriving smaller-sized graphs while retaining the important patterns learned by the GNN model. Graph pruning enhances capability of GIN-Graph in generating fine-grained explanation graphs of GNNs, allowing it to interpret GNNs at varying levels of granularity.

\section{Conclusion}
In this paper, we investigate the properties of model-level explanation graphs, and define a validation score and a granularity related explanation metric to evaluate explanation graphs. We also proposed GIN-Graph to generate high quality model-level explanation graphs that are similar to real graphs, meanwhile maximizing the prediction probability for a certain class. Experimental results demonstrate that GIN-Graph presents superior performance to GNNI for generating valid explanation graphs with better stability and reliability, highlighting its generality in generating model-level explanation graphs for GNNs trained on a variety of graph datasets. In the future, given the considerable diversity within models learned on different datasets, we aim to tackle the challenge of creating a universal metric for accurately evaluating explanation graphs across all models. Additionally, we plan to focus on reducing the reliance on human expertise and prior knowledge in identifying meaningful explanation graphs.

\bibliography{references}
\end{document}